# Mathematics Teachers' Interactions with a Multi-Agent System for Personalized Problem Generation

Candace Walkington[1], Theodora Beauchamp[1], Fareya Ikram[2], Merve Koçyiğit Gürbüz[1], Fangli Xia[1], Morgan Lee[3] and Andrew Lan[2]

[1]Southern Methodist University, Dallas TX 75205, USA
cwalkington@smu.edu, theob@smu.edu, mkocyigitgurbuz@smu.edu, fanglix@smu.edu

[2] University of Massachusetts Amherst, Amherst MA 01003, USA
fikramn@umass.edu, andrewlan@umass.edu

[3] Worcester Polytechnic Institute, Worcester, MA 01609 USA
mplee@wpi.edu

**Abstract.** Large language models can increasingly adapt educational tasks to learners' characteristics. In the present study, we examine a multi-agent teacher-in-the-loop system for personalizing middle school math problems. The teacher enters a base problem and desired topic, the LLM generates the problem, and then four AI agents evaluate the problem using criteria that each specializes in (mathematical accuracy, authenticity, readability, and realism). Eight middle school mathematics teachers created 212 problems in ASSISTments using the system and assigned these problems to their students. We find that both teachers and students wanted to modify the fine-grained personalized elements of the real-world context of the problems, signaling issues with authenticity and fit. Although the agents detected many issues with realism as the problems were being written, there were few realism issues noted by teachers and students in the final versions. Issues with readability and mathematical hallucinations were also somewhat rare. Implications for multi-agent systems for personalization that support teacher control are given.

## 1 Background

### 1.1 Introduction

Generative AI offers the promise of a personalized learning experience for each student that connects to their individual characteristics [1]. At the K-12 level, students are often highly interested in popular culture and social activities like movies, music, video games, and sports. There is potential for Large Language Models (LLMs) to provide a personalized mathematics learning experience. However, when considering K-12 students, there are serious concerns about LLMs being directly student-facing, with no teacher-in-the-loop to engage with LLM-generated content. LLMs may generate harmful or misleading content [2, 3] or may generate content that lacks pedagogical appropriateness [4] or that does not reflect the deep knowledge that teachers typically have about their students [5]. There are also concerns that student-facing LLMs may create distance between teachers and students, compromising the relationships they form [6]. Indeed, when teachers are involved in decisions about personalizing learning for students, they bring in rich professional knowledge of their students' needs and of their curriculum and can learn more about their students and their interests, growing their professional knowledge [7]. For these reasons, we have been exploring teacher-in-the-loop approaches to context personalization. Here we examine teacher prompting moves for and students' reactions to LLM-generated personalized problems. Our research question is: What are the affordances and limitations of a multi-agent LLM approach to generating personalized tasks? This paper contributes: (1) a multi-agent, teacher-in-the-loop architecture for context personalization, (2) evidence from authentic classroom use on which dimensions of personalization are hardest to achieve, and (3) design implications for balancing teacher control with student-level authenticity.

### 1.2 Theoretical Framework

Interest theory [8] describes how tasks that are relevant can trigger and maintain students' interest in learning, leading to increased attention, engagement, and use of learning strategies, which in turn can lead to increased learning and interest in the academic area. [9] show how this interest development process unfolds for personalized algebra tasks, demonstrating that these tasks trigger situational interest in the task which progresses into increased efficiency, increased learning, and increased interest in mathematics as a subject area. Personalized tasks may also allow students to draw upon their prior knowledge of pursuing their interests, providing contextual grounding [10].

However, not all studies that have explored context personalized have found positive results (e.g., [11]). Mathematics story problems used in K-12 settings often lack authenticity and realism [12] – they describe activities and events that would never reasonably happen in the real world with stories that are contrived simply to make the math "work." [13] cite the *depth* of personalized tasks, in terms of their match to actual real-world activities of the solver, as a key component of effectiveness. Further, readability demands can hamper mathematical word-problem performance because linguistic complexity draws on limited working-memory resources that are required to construct a coherent situation model [14]. Particular readability features can predict instances where weaker readers will struggle [15]. Further, when LLMs generate mathematics problems, there is still evidence that they can create problems that are unsolvable, that use incorrect units, or that have math errors or inconsistencies [16]. Thus, when personalizing problems, it is important the LLMs include components that minimize readability issues and hallucinations while maintaining authenticity and realism.

### 1.3 Literature Review

Prior research has examined the effects of using LLMs to create personalized math problems. [1] examined elementary school students receiving LLM personalized math word problems, finding that these problems enhanced intrinsic motivation, interest, and performance. This study's approach involved the system creating personalized problems for students on-the-fly without a human in-the-loop. In [16], 3rd-5th grade students were given problems that were were LLM-generated based on general topics like sports or video games. Problems were screened by teachers, and student performance was compared to human-generated problems. Students performed similarly on AI-generated and human-generated problems, but preferred the LLM-generated problems. [17] investigated 7[th] grade math teachers personalizing problems with ChatGPT, examining teacher prompts and student responses. They found that teachers spent a lot of time and prompts adjusting the realism of the problems generated by ChatGPT, including the numbers, quantities, and other elements of the context. Students often commented on the fit of the problems to their particular interests and rated both the interestingness of the problems and the closeness of the problems to their interests relatively low. Finally, some research suggests that LLM-generated content may increase reading burdens on students [16]. Based on these theories and literature, we developed a multi-agent system for teacher-in-the-loop context personalization that evaluates authenticity, realism, readability, and mathematical accuracy. We next describe testing this system with students and teachers.

## 2   Method

### 2.1   Participants

Participants included 8 7[th] grade mathematics teachers in the United States, who were recruited through ASSISTments' [18] teacher mailing list. The teachers (2 male, 6 female) had an average of 14.8 years of experience, with 7 teachers identifying as White and 1 as Black. Seven teachers taught in public schools, and 1 taught at a private school. In terms of 7[th] grade students, the study used an opt-out parental consent method for some data sources (1-5 ratings and problem-solving actions), and an opt-in parental consent method for other data sources (responses to open-ended and demographic items). We had 9 students whose parents opted out, and 364 whose parents did not opt out. We had 207 students whose parents opted them into the research, and therefore we were able to collect demographic data from them. Of these students, 105 identified as female, 95 as male, and 7 as Other/Prefer not to Say. One hundred and sixty-three identified as White, 31 as Hispanic, 23 as Black, 9 as Native American/Pacific Islander, 11 as Asian, and 15 as Other.

### 2.2   Persona Problem Builder Environment

The Persona Problem Builder (PPB) is a web-based platform developed using React for frontend code and Flask API for the backend. GPT-4o underlies the system, leveraging its math reasoning capabilities. The application is hosted on an EC2 instance. The PPB is designed to integrate with ASSISTments, an online homework system [18]. A teacher provides an ASSISTments problem ID and a topic (e.g., baseball) and can choose whether to retain or modify the original numerical values in the problem. The system then generates a personalized version of the problem. The initial problem is produced using a structured prompt that targets topic alignment and preservation of the underlying mathematics. The generated problem is then evaluated by four agents:

- Authenticity agent: Evaluates whether the context is age-appropriate and likely to be relatable for middle school learners.
- Realism agent: Evaluates whether quantities, units, and contextual details are plausible for the chosen topic (e.g., whether scores or time intervals fall within reasonable ranges).
- Reading-level agent: Evaluates whether vocabulary and sentence complexity match the intended reading level.
- Hallucination agent: Evaluates whether the problem is mathematically consistent with the original, e.g., whether there is a correct option in multiple-choice items.

Each agent has a specified role and prompt and each returns a binary pass/fail decision. For failed cases, the agent returns a set of identified issues. These issues are combined and provided to a refinement agent that revises the problem. The process iterates until all agent evaluations pass or a maximum number of refinement steps is reached (five in this study). After refinement, teachers can engage with further authoring the problem using the interface and then submit it to ASSISTments to be built.

### 2.3   Data Sources and Analysis

Each teacher selected 4-8 practice word problems in 1 unit of the *Illustrative Mathematics* (IM) open-source $7^{th}$ grade curriculum. Then, the teacher was asked to use the PPB to write 4 different versions of each problem that fit with three different interest topics (e.g., sports, video games). The teachers had all previously distributed interests surveys to their students. The first data source was the text of the 212 personalized problems that teachers wrote using the PPB.

The second data source was logs of the agents' problem-building actions in response to the teachers' initial requests to generate personalized problems. The teachers made 212 such problem-generation requests that were responded to by the 4 agents. The third data source was the epistemic moves that the teachers made to further edit and refine the 212 problems the agent teams generated. There were 179 such teacher moves made for 46 problems; 166 problems were accepted as-is, with no further prompting needed.

The fourth data source was students' open-ended feedback on the personalized problems. The following 3 questions were posed to students for 47 different personalized mathematics problems: (1) Was the last problem interesting to you? Why or why not? (2) How do you think the last problem could be improved to better match with things you are interested in? (3) Was there anything in the last problem that was incorrect relating to your interest area and how it works?. There were 237 students with parental consent who responded to these questions.

Data was compiled from logs of teachers' interactions with the PPB. We looked at teacher behaviors as they prompted the PPB to refine the problems. We coded the kinds of prompts the teachers used using thematic analysis [19], where the unit of analysis was one teacher prompt, and codes were compiled under themes related to each agent. We then used similar techniques to code students' open-ended feedback, considering the three questions together as one coding unit. Finally, we calculated descriptive statistics of readability measures using Coh-Metrix [20].

## 3 Results

### 3.1 Evaluation of Authenticity Agent

Of the 179 moves teachers used to refine the 46 personalized problems, several codes related to the authenticity theme. In 48 of the 179 moves, the code indicated that the teacher adjusted the problem's topic; e.g., "Add that the student loves Dr. Pepper but his parents won't buy for him." or "change pop to hiphop" or "Could you reframe this question around the band selling cheesecakes for their fundraiser?" There were 16 prompts coded where teachers asked the LLM to adjust the problem to fit their local geographical context and 17 moves coded where the teacher asked the LLM to change names of people, often to the names of students in their class.

Students' feedback revealed further insights. A total of 235 student responses (out of 422) were coded as commenting on the problem's topic, and 83 of those responses were coded as the student saying the topic of the problem was something that they liked. Some examples included: "it was about Taylor swift and I was listening to one of her songs at that exact moment and it was perfect" and "Yes it was interesting because it was about *Call of Duty* and I like *Call of Duty*." However, 160 responses, almost twice as many, described how the topic of the problem was something the students did not like or did not engage with in their lives. Some examples included: "The last math problem wasn't interesting to me, mainly because I hate Taylor Swift, not as a person but as her music to me in my opinion low-key sucks" and "Meh, *Call of Duty* isn't really MY thing." As can be seen from these quotes, what is authentic to one student may not be authentic to another. Only 4 student responses (out of the 422 response sets) pointed out more global issues with authenticity when giving feedback on the problems, with comments like "No kids play that anymore" and "I literally don't think about circumference every time I walk around an area shaped like a circle!".

An additional 9 student responses (out of the 422 response sets) were coded as "meta" comments critiquing how the personalization was being implemented. Some comments went against our broad grain size, teacher-in-the-loop approach to personalization. One student described how "I think we should be able to put in things we are interested in so that you guys know what our interests are. :D" while another said "I don't know if this is possible, but maybe include multiple artists so multiple people have who they like, or certain problems can be assigned to certain students with interest in certain artists." Other students gave advice on how to make better topic choices, with comments like "It could be something directed to a bigger fanbase, then you will most likely get more people to like it." Still other students objected to the idea of personalization itself, stating that "I don't want to like math, this feels awful." This quote makes the point that personalization is not universally desirable and also raises the question of whether having a teacher-in-the-loop to be part of the personalization makes it more likely to be better received by students, versus the students believing that a disembodied AI is trying to appeal to their interests in shallow ways.

### 3.2     Evaluation of Realism Agent

There were 10 instances coded under the "realism" theme, where the teacher explicitly let the LLM know that the problem it had created was not realistic – with prompts like "Can you have fractional blocks in Roblox, and can you buy them with Robux?" and "We can't have them eat 7.5 boxes. People don't eat half of a box of mac and cheese." In addition, there were 20 prompting moves coded under the realism theme where the teacher asked the LLM to modify the quantity or unit in the problem, with prompts like "change penalty to fumbles" and "change gems back to dollars." For these prompts, it is less clear if teachers' modification were due to realism issues, but there was some motivation for the teacher to want to adjust the kinds of objects and measurement units cited in the problem. In addition, there were 6 student open-ended comments coded under the realism theme where students flagged additional issues with the realism of the final personalized problems. These included comments like "In the game, there is no tax on items" and "Unless they're really good at making sushi, 8 rolls a minute is kinda absurd."

### 3.3     Evaluation of Readability Agent

We next examined how the original problems from the IM curriculum that were given to the PPB (of which there were 32 unique problems) compared in terms of readability to the 156 final versions of the personalized problems that were given to students. This comparison is shown in Table 1. As can be seen from the table, the PPB maintained relatively comparable reading levels on three of the four measures. The PPB did appear to increase Flesh-Kincaid to be more aligned with the grade level (grade 7) of the students. This may have been because the LLM was told that these were 7[th] grade students.

Table 1. Readability metrics on original and LLM-generated problems

| Readability Measure | Avg. for Original (SD) | Avg. for Personalized (SD) |
| --- | --- | --- |
| Flesch-Kincaid | 4.38 (2.47) | 6.62 (2.25) |
| Word Count | 48.25 (23.89) | 49.85 (16.87) |
| Word Concreteness | 404.67 (43.97) | 401.35 (40.12) |
| Second Person Pronouns | 6.06 (12.48) | 6.27 (18.72) |

There were only 7 prompts where teachers asked the LLM to adjust readability – some examples were "Make this reading level 7th grade" and "make this less wordy." There were 8 student responses from students where students mentioned issues with readability. Some examples were: "Maybe making it easier to understand in the writing of the question" and "I did not love the format of the question because it really confused me." There were also 10 comments from students that mentioned that they wanted the problem contexts to be longer and more detailed; such comments included "It could maybe include more storytelling. that would definitely help encourage the person to work on it more" and "The last problem was not interesting because it didn't give enough detail." These comments suggest that some of the students wanted problems that were technically less readable, in order to allow the problems to be more contextually interesting to them.

### 3.4 Evaluation of Mathematical Hallucination Agent

We did not see any instances of teachers identifying mathematical errors in the final version of the problem. However, teachers did have 25 prompt moves where they either requested mathematical clarification be made to part of the story context or requested an adjustment of the mathematical vocabulary being used in the story context. Some examples were "can you use the word credit in the problem," "please make it about circles" (this was for a problem that was originally about circles, which got modified into a generic proportion problem without circles in it), and "lets change the variable $g$ to $d$." The teacher also critiqued the number choice made by the LLM in 16 cases, with prompts like "change 90 percent to 75 percent" and "change the $6 to $6.67" This shows that although actual errors are rare, teachers' professional mathematical knowledge is still used in potentially significant ways to adjust the mathematical properties of LLM-generated problems. There were additionally 18 comments from students that described the lack of clarity of the mathematics in the personalized problems, with comments including "I couldn't figure out if it was a positive or negative" and "the problem itself didn't make sense."

## 4    Discussion

### 4.1   Attaining Authenticity and Realism in Multi-Agent Systems

The biggest challenge we encountered with our design was attaining authenticity of the mathematical tasks. The researchers and the teachers wanted the tasks to reflect the ways in which students engage or might reasonably engage with their interest areas in their day-to-day life [12]. However, we found that this was very hard to accomplish with teacher-in-the-loop LLM-generated personalization because students' interests were so variable and fine-grained. While a small grain size does not guarantee an authentic problems, it may often be a necessary but not sufficient condition for authenticity to occur. A model with the teacher out-of-the-loop has been used in other studies [1]; this level of granularity, where problems are created on-the-fly for individual students, was what students expected. We also saw that personalization may elicit resistance when students perceive it as externally imposed rather than self-authored [13].

A key purpose of using the personalized problems is to activate interest [8,13], and the granularity issue might have compromised this goal, as it may have caused there to be less activation of *personal relevance*. [16] used a similar broad-grain-size approach but did not describe any issues with granularity in their results. This may be because both open-ended feedback from students and iterations of refinement with teachers need to be built into personalization processes for such issues to be detected.

Despite this, we saw that a teacher-in-the-loop model had advantages – teachers could enhance mathematical clarity, readability, and realism, and there was one case of a teacher flagging inappropriate content. This relates to other work suggesting that when teachers modify LLM output to suit their needs and context - rather than just giving LLM-generated tasks to students as-is – students like the tasks better [17].

Our approach to ensuring realism worked far better than authenticity, and it evidences how susceptible uninstructed LLMs are to make unrealistic problems. Lack of realism can limit students activating their prior knowledge [10, 12, 13]. While there are no clear experimental studies that isolate the effect of lack of realism, this issue can compromise bigger-picture learning goals for students to see mathematics as sensible, meaningful, and relevant.

### 4.2   Mathematical Hallucinations and Readability

There were few issues with mathematical hallucinations or readability. Teachers changed to the mathematical clarity, mathematical vocabulary, and readability of the problems before they went to students. While it is well-documented that readability matters for student performance [14,15], it is less clear the degree to which hallucinations like unclear problem statements negatively impact students. Overall, contemporary LLMs have advanced considerably in addressing readability issues. Having a teacher in-the-loop may be important for issues with mathematics language and clarity.

### 4.3   Recommendations and Next Steps

Our study reveals ways in which multi-agent systems can support teachers in personalizing mathematics problems. In current work, we have moved our system to the more capable GPT-5.2 model, while also using functionality that allows teachers to set the weights of the different autonomous agents. For example, if the teacher is not concerned about reading level, they can turn down the weight of the readability agent. Better approaches are still needed to allow for problem authenticity. If the teacher was taken out-of-the-loop, considerable work would need to be put into safety and contextual fit, as well as to the ways in which the language and the math are presented to students. And still, this would not be ideal - teachers were not merely correcting surface errors as they engaged with the PPB; their interventions reflected deep pedagogical knowledge about students' lived experiences, linguistic appropriateness, and mathematical framing.

**Acknowledgments.** This work was supported by the National Science Foundation under Grant DRL 2341948. Any opinions, findings, conclusions, or recommendations expressed in this material are those of the author(s) and do not necessarily reflect the views of the National Science Foundation.

**Disclosure of Interests.** The authors have no competing interests to declare that are relevant to the content of this article.

Multi-Agent Personalized Problem Generation     9